\documentclass{article} 
\usepackage{iclr2026_conference,times}

\usepackage{amsmath,amsfonts,bm}

\def\eqref#1{equation~\ref{#1}}

\def\1{\bm{1}}

\def\vc{{\bm{c}}}

\def\vp{{\bm{p}}}

\def\vx{{\bm{x}}}

\def\vz{{\bm{z}}}

\DeclareMathAlphabet{\mathsfit}{\encodingdefault}{\sfdefault}{m}{sl}
\SetMathAlphabet{\mathsfit}{bold}{\encodingdefault}{\sfdefault}{bx}{n}

\usepackage[utf8]{inputenc} 
\usepackage[T1]{fontenc}    
\usepackage{hyperref}       
\usepackage{url}            
\usepackage{booktabs}       
\usepackage{amsfonts}       
\usepackage{nicefrac}       
\usepackage{microtype}      
\usepackage{xcolor}         
\usepackage{subcaption}
\usepackage{graphicx}
\usepackage{natbib}
\usepackage{amsmath}
\usepackage[noabbrev]{cleveref}
\usepackage{algorithm}
\usepackage{algorithmic}
\usepackage{ulem}
\usepackage{multirow}
\usepackage{wrapfig}
\usepackage{comment}
\usepackage{siunitx}

\crefname{appendix}{Appendix}{Appendices}
\Crefname{appendix}{Appendix}{Appendices}

\title{Post-hoc Stochastic Concept Bottleneck Models}

\author{
  Wiktor Jan Hoffmann, Sonia Laguna\thanks{Correspondence to \href{mailto:slaguna@inf.ethz.ch}{\texttt{slaguna@inf.ethz.ch}}}, Moritz Vandenhirtz, Emanuele Palumbo, Julia E.~Vogt \\
  Department of Computer Science, ETH Zurich, Switzerland \\
}

\iclrfinalcopy 
\begin{document}

\maketitle

\begin{abstract}
Concept Bottleneck Models (CBMs) are interpretable models that predict the target variable through high-level human-understandable concepts, allowing users to intervene on mispredicted concepts to adjust the final output. While recent work has shown that modeling dependencies between concepts can improve CBM performance, especially under interventions, such approaches typically require retraining the entire model, which may be infeasible when access to the original data or compute is limited. In this paper, we introduce Post-hoc Stochastic Concept Bottleneck Models (PSCBMs), a lightweight method that augments any pre-trained CBM with a multivariate normal distribution over concepts by adding only a small covariance-prediction module, without modifying the backbone model. We propose two training strategies and show on real-world data that PSCBMs consistently match or improve both concept and target accuracy over standard CBMs at test time. Furthermore, we show that due to the modeling of concept dependencies, PSCBMs perform much better than CBMs under interventions, while remaining far more efficient than retraining a similar stochastic model from scratch.
\end{abstract}

\section{Introduction}
\label{sec:introduction}
The adoption of machine learning models in high-stakes domains is often limited by their lack of interpretability due to their black-box nature \citep{lipton2016mythos,doshi-velez2017towards}.
Concept Bottleneck Models (CBMs), first introduced by \citet{koh2020concept}, address this by inserting a layer of neurons aligned with human-understandable concepts before the target prediction.
A CBM is composed of a concept encoder that predicts the concepts, and a prediction head that takes these concepts as inputs; the final prediction is thus explained through the intermediate concept values. While the original formulation assumes independence among concepts, accounting for correlations has been shown to improve the model's performance \citep{havasi2022addressing,singhi2024improvinginterventionefficacyconcept,Vandenhirtz2024stochastic}. However, existing approaches that capture concept dependencies typically require training the entire model with dedicated objectives. In contrast, we demonstrate that such dependencies can be incorporated post-hoc into a pre-trained CBM. A more detailed discussion of related work is provided in~\Cref{app:related-work}. 

Building on this idea, we introduce Post-Hoc Stochastic Concept Bottleneck Models (PSCBMs)\footnote{The code is available at \url{https://github.com/wiktorhof/PSCBM}.}, which extend pre-trained CBMs with a lightweight module that predicts concept covariance.
Inspired by Stochastic Concept Bottleneck Models (SCBMs)~\citep{Vandenhirtz2024stochastic}, we model concept dependencies with a multivariate normal distribution. 
This approach not only improves predictive performance, but also enables more efficient concept interventions—the process by which a user modifies predicted concepts to directly influence downstream predictions. 

This work contributes to model interpretability through human-understandable concepts in three ways: \textit{(i)} We introduce PSCBM, a lightweight post-hoc module that incorporates concept correlations into existing CBMs without requiring full retraining, thereby reducing compute requirements. \textit{(ii)} We propose a simple intervention-based training procedure that further improves intervention efficiency in (P)SCBM-like models. \textit{(iii)} We show on real-world data that PSCBM improves both predictive accuracy and intervention effectiveness while remaining substantially more efficient than full model retraining.

\section{Method}
\vspace*{-0.2cm}

\label{sec:methods}

\begin{figure}
    \centering
        \includegraphics[width=0.9\linewidth]{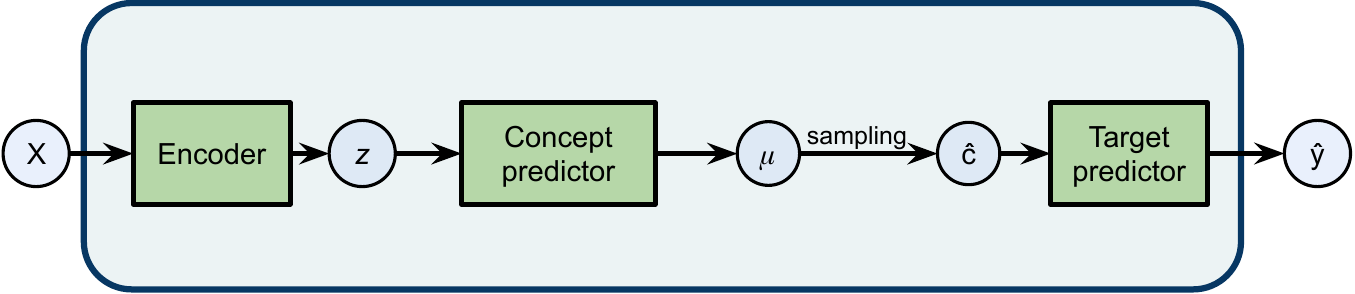}
        \mbox{}\\
        \includegraphics[width=0.9\linewidth]{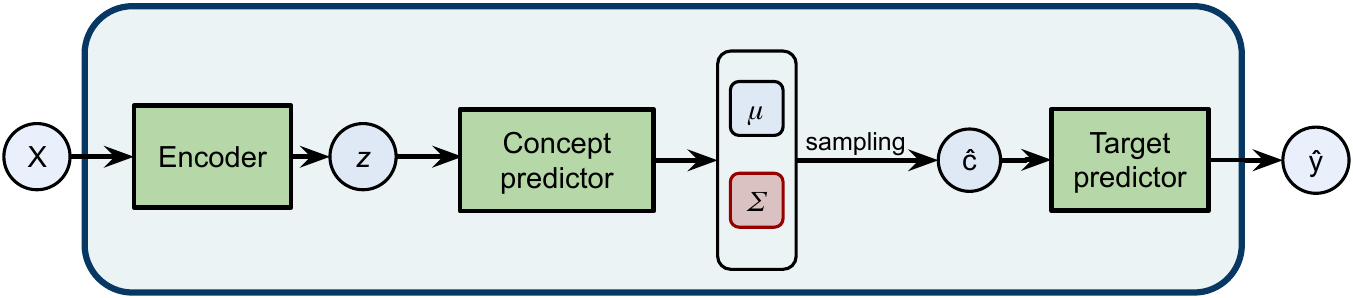}
    \caption{Comparison of a hard CBM (top), SCBM (bottom), and our proposed PSCBM (bottom, red block). All methods input $\vx$ to an encoder to produce the feature vector $\vz$ from which concepts are obtained and fed to the target predictor. CBM directly predicts concept values $\hat{\vc}$, while in SCBM, the predictor outputs an expected value $\boldsymbol{\mu}$ and a covariance matrix $\boldsymbol{\Sigma}$ that define a multivariate normal distribution to sample from. PSCBM incorporates the $\boldsymbol{\Sigma}$ predictor (red box) to a frozen pre-trained CBM.\looseness-1}
    \label{fig:model-schemes}
\end{figure}

\paragraph{Modeling concept dependencies with a normal distribution}
\label{sec-modelling-dependencies} 

In this paper we propose PSCBMs, a post-hoc extension of CBMs that enables to model dependencies between concepts without retraining the model from scratch. A CBM \citet{koh2020concept} consists of three main components: a feature encoder, a concept predictor, and a target predictor. The feature encoder $h$ (e.g., a CNN or ResNet) extracts features $\vz = h(\vx)$ from an input $\vx$. Then the concept predictor $g$  outputs concept probabilities $\vp = g(\vz)$. When focusing on binary concepts for downstream prediction (i.e. \textit{hard} CBMs~\citep{havasi2022addressing}), rather than their probabilities (\textit{soft} CBMs), the next step is sampling $M$ concept values from a Bernoulli distrbution $\hat{\vc}^{(1)}, \dots \hat{\vc}^{(M)} \sim \text{Bernoulli}(\vp)$. Then the target predictor $f$ maps the sampled concepts to the final prediction $\hat{y} = \frac{1}{M}\sum_{m=1}^M f(\hat{\vc}^{(m)})$. SCBMs \citep{Vandenhirtz2024stochastic} extend this framework by explicitly capturing correlations between concepts: the concept predictor is extended to feature a mean predictor $g_\mu$ and a covariance predictor $g_\Sigma$, which define a multivariate normal distribution that models concept dependencies. Concepts are sampled as $
\hat{\vc}^{(m)} = \sigma(\eta^{(m)})$ for $m = 1, \dots, M$, where $\eta^{(m)} \sim \mathcal{N}(\boldsymbol{\mu}, \boldsymbol{\Sigma})$ and $\sigma$ is the sigmoid function. Our proposed PSCBMs directly build upon this idea: given a pre-trained CBM, we reuse its concept predictor $g$ as $g_\mu$ and add a lightweight covariance predictor $g_\Sigma$, training only the latter while keeping the rest of the model frozen. In this way, any existing CBM can be turned into a stochastic, dependency-aware model post-hoc, as illustrated in~\Cref{fig:model-schemes}. 

\paragraph{Interventions in Stochastic Concept Bottleneck Models}
A distinguishing property of CBMs is that they allow users to modify predicted concept values at test time and thereby directly update the downstream prediction. This process, known as intervention, can be factorized into two parts: the policy $\pi$, which selects the concepts to be intervened on (e.g., randomly or based on uncertainty), and the strategy $\tau$, which determines how their values are updated. We describe several policies and strategies in~\Cref{app-intervention-strategies}.
Intervening in SCBMs or PSCBMs additionally accounts for concept dependencies and proceeds in four steps:
\begin{enumerate}
    \item \textbf{Select concepts for intervention} Use a policy $\pi$ to choose a subset $\mathcal{S}$ of concepts.
    \item \textbf{Set intervened logits} Assign values to $\boldsymbol{\eta}'_{\mathcal{S}}$, the logits of intervened concepts, according to a chosen intervention strategy $\tau$.
    \item \textbf{Update remaining logits} Compute the conditional normal distribution parameterized by $\boldsymbol{\bar\mu}, \boldsymbol{\bar\Sigma}$ for the non-intervened concept set $\setminus \mathcal{S}$, using the equations for a conditional normal distribution:
    $\boldsymbol{\eta}_{\setminus \mathcal{S}} \mid \vx, \boldsymbol{\eta}'_{\mathcal{S}} \sim \mathcal{N}\left(\boldsymbol{\bar\mu}, \boldsymbol{\overline{\Sigma}}\right)$, 

    where
   $\boldsymbol{\bar\mu} = \boldsymbol{\mu}_{\setminus \mathcal{S}} + \boldsymbol{\Sigma}_{\setminus \mathcal{S},\mathcal{S}} \boldsymbol{\Sigma}_{\mathcal{S},\mathcal{S}}^{-1}(\boldsymbol{\eta}'_{\mathcal{S}} - \boldsymbol{\mu}_{\mathcal{S}}),
   \boldsymbol{\overline{\Sigma}} = \boldsymbol{\Sigma}_{\setminus \mathcal{S},\setminus \mathcal{S}} - \boldsymbol{\Sigma}_{\setminus \mathcal{S},\mathcal{S}} \boldsymbol{\Sigma}_{\mathcal{S},\mathcal{S}}^{-1} \boldsymbol{\Sigma}_{\mathcal{S},\setminus \mathcal{S}}.$
    \item \textbf{Sample probabilities} Sample binary concept values ${\vc}_{\setminus \mathcal{S}}$  from the logits $\boldsymbol{\eta}_{\setminus \mathcal{S}}$.
\end{enumerate}

\paragraph{Learning the covariance matrix post-hoc}
\label{sec-learning-post-hoc}
The covariance predictor $g_{\boldsymbol{\Sigma}}$ is trained by minimizing the loss function of a regular SCBM. It teaches the model to predict both the concepts and target variable correctly and encourages sparsity in the covariance matrix:
\vspace{-0.12cm}
\begin{equation*}
\label{eq-scbm-loss}
    \mathcal{L} = \underbrace{ -\log \sum_{m=1}^M \exp \sum_{i=1}^C -\mathrm{BCE}\left(c_i, \sigma (\eta_i^{(m)})\right)}_{\text{Concept Loss}} + \lambda_1\underbrace{\mathrm{CE}\left( y, 
    \frac{1}{M} \sum_{m=1}^M g_{\boldsymbol{\psi}}(\vc^{(m)})
\right)}_{\text{Target Loss}} + \lambda_2\underbrace{\sum_{i\neq j} \boldsymbol{\Sigma}^{-1}_{i,j}}_{\text{Regularization}}.
\end{equation*}
Here, $M$ is the number of Monte Carlo concept samples, $C$ is the number of concepts, $\text{BCE}$ and $\text{CE}$ denote the (Binary) Cross-Entropy, $\vx$ is the input, $y$ the true label, $c_i$ the true concept values, $\eta^{(m)}$ 
the sampled logit vectors,
$\sigma(\cdot)$ the sigmoid function, $\vc^{(m)}$ the binary concept values sampled from the Bernoulli distribution defined by $\sigma(\eta^{(m)})$, $\boldsymbol{\Sigma}(\vx)$ the predicted covariance matrix and $g(\cdot)$ the target predictor. Finally, $\lambda_1$ and $\lambda_2$ are weighting parameters regulating the loss strength.

We propose two training paradigms for PSCBM, which can be extended to SCBM: \textit{(i)} The loss function is applied to the model's predictions without concept interventions. \textit{(ii)} We encourage more responsive predictors to concept interventions. In every training iteration, for every sample in the training dataset, we randomly select a set $\mathcal{S}$ of concepts to intervene on using some strategy $\tau$ and calculate the loss after the intervention. The cardinality of $\mathcal{S}$ is fixed throughout training. This is done $N$ times for each data point and the average of the loss for all these interventions is calculated. This method is similar to \textit{RandInt} introduced by \citet{espinosa2022concept} but it differs in two ways: \textit{(i)} Unlike \textit{RandInt}, which decides independently for each concept whether to replace it with its true value, our method enforces that a fixed number of concepts is intervened on in every case. \textit{(ii)} At each training iteration, multiple random interventions per sample are made and the loss is averaged. This reduces gradient variance and yields more stable training compared to applying a single intervention. A detailed pseudocode for the calculation of this loss can be found in \Cref{app-pseudocode}.

\section{Results}
\label{sec-results}
In our experiments, we address two main questions: \textbf{1.} How does post-hoc learning of the covariance affect the test accuracy without interventions? \textbf{2.} How does it affect predictions after interventions? 
%
\paragraph{Experimental setup}
We compare the performance of PSCBM to SCBM and regular CBM. PSCBM indicates training without concept interventions, while PSCBMi includes interventions during training.
SCBM proposes two ways to learn the covariance matrix, amortized per instance, $\boldsymbol{\Sigma}(\vx)$, or learnt globally, $\boldsymbol{\Sigma}$. For brevity, we present the results on the global covariance, and include their amortized counterparts in \Cref{app-amortized-covariance}. All implementation details are described in \Cref{app-implementation}. For test-time interventions, concepts are selected based on their associated prediction uncertainty, i.e., the concept with predicted probability closest to 0.5 is chosen. Unlike CBM where the intervened probabilities are set to 0/1, SCBMs and PSCBMs set intervened concept logits to those corresponding to $\epsilon$ or $1-\epsilon$, respectively, where $\varepsilon$ is small for stability. We evaluate our methods on the Caltech-UCSD Birds-200-2011 dataset \citep{wah2011caltech}, composed of photographs of birds of 200 different classes. We use the variant introduced in \citet{koh2020concept} for CBMs with 112 binary per-class concepts and, as evaluation metrics, we focus on concept and target accuracy.

\vspace{-0.1cm}
\paragraph{Test performance}
\label{sec-test-performance}
In \Cref{tab-test-accuracy} we present concept and target accuracy of the evaluated models without interventions. The results show that adding our covariance learning module improves performance. Both variants of PSCBM outperform SCBM on both concept and target accuracy. For target accuracy, PSCBM trained without interventions clearly surpasses the regular CBM, while on concept accuracy the two achieve comparable results. 
The training times presented in the last column show that training a PSCBM without interventions is computationally inexpensive due to the fact that only one module is trained. Training with interventions however requires significantly more time.
\begin{table}[h!]
    \centering
    \caption{Test-time target and concept accuracy without interventions and their AUCs under interventions. AUC is normalised to stay within the interval [0,1]. We report the mean and standard deviation over three runs for baselines and nine runs for PSCBM. The best scores for each metric are highlighted with \textbf{bold} font, and those within a standard deviation from the best one are \underline{underlined}.}
    \label{tab-test-accuracy}
    \resizebox{\textwidth}{!}{%
    \begin{tabular}{@{}lccccc@{}}
    \toprule
    \textbf{Method}         & \textbf{Target Accuracy (\%)} & \textbf{Concept Accuracy (\%)} & \textbf{Target Accuracy AUC} & \textbf{Concept Accuracy AUC}  & \textbf{Training time (s)} \\
    CBM                     & $ 67.3973	\pm 0.5722 $ &	$ \mathbf{94.9403}	\pm 0.1059 $ 			& $ 0.9551 \pm 0.0000 $ & $ 0.9825 \pm 0.0000 $  & 
    $   7204  \pm        247  $ \\
    SCBM                   & $ 65.5173	\pm 0.8539 $ &	$ 94.4569	\pm 0.1328 $ & $ 0.9671 \pm 0.0001 $ & $ \mathbf{0.9870} \pm 0.0000 $  & 
     $   8134  \pm        767 $  \\
    PSCBM                  & $ \mathbf{68.3983}	\pm 0.1992 $ &	$ \underline{94.9285}	\pm 0.0206 $& $ 0.9680 \pm 0.0003 $ & $ 0.9859 \pm 0.0001 $ & 
    $    740  \pm         94  $  \\
    PSCBMi               & $ \underline{68.1970}	\pm 0.1274 $ &	$ \underline{94.9026}	\pm 0.0350 $ & $ \mathbf{0.9704} \pm 0.0002 $ & $ 0.9866 \pm 0.0001 $ & $  14084  \pm        267 $ \\
    \bottomrule
    \end{tabular}}
\end{table}

\paragraph{Intervention performance}
\label{sec-intervention-performance}

\begin{figure}[h!]
    \centering
    \includegraphics[width=\linewidth]{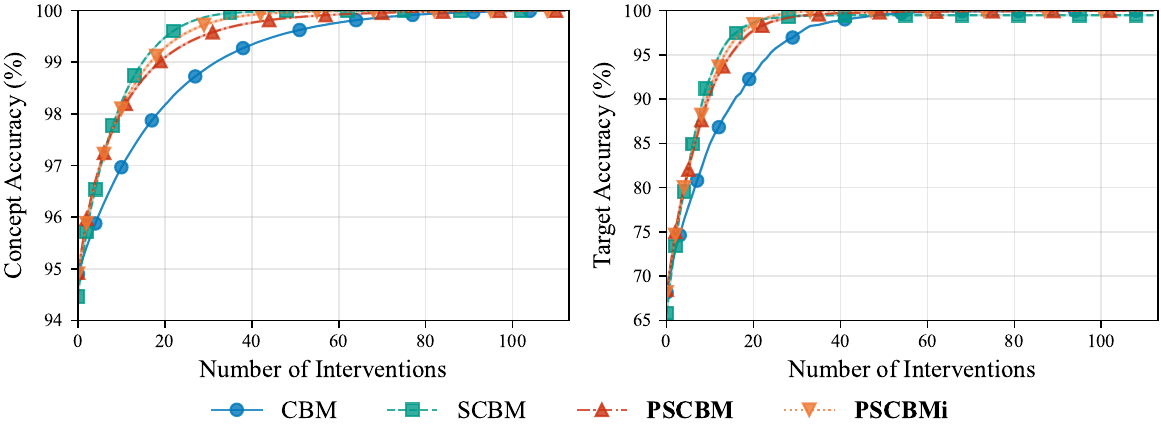}
    \caption{Intervention curves of Concept and Target Accuracy for PSCBM models and baselines when concept uncertainty policy is used. Confidence intervals are thinner than the lines.}
    \label{fig:intervention-curves}
    \vspace{-0.3cm}
\end{figure}

To evaluate intervention performance we rely on two approaches: visual inspection of the intervention curves, and the area under the intervention curve (AUC) \citep{singhi2024improvinginterventionefficacyconcept}, which summarizes each curve into a single value for easier comparison. The last two columns of \Cref{tab-test-accuracy} report the AUC scores. SCBM achieves the highest concept AUC, but both PSCBM variants outperform it on target AUC, with PSCBMi being the strongest overall. This highlights the benefit of training with interventions. Both PSCBM models also outperform the regular CBM on both metrics, showing that adding a covariance matrix post-hoc improves intervention performance. 
For a more detailed comparison, we examine the intervention trajectories under the uncertainty policy in \Cref{fig:intervention-curves}. We additionally include the corresponding results when selecting concepts with a random policy in~\Cref{app-random-policy}. Both PSCBM variants consistently outperform CBM, with the intervention-trained variant performing best. However, for a few interventions, PSCBM without interventions during training lags behind SCBM, indicating that joint training of the covariance is more effective. PSCBM with interventions surpasses SCBM on target accuracy after about 20 interventions, though it does not reach SCBM’s performance on concept accuracy. We note that SCBM could also be trained with interventions, which might further boost its results. 
It is clear from these that adding a covariance on a trained CBM can significantly improve its performance.

\section{Conclusion}
In this paper, we introduce PSCBM, a lightweight extension to CBMs that models concept dependencies via a multivariate normal distribution. Crucially, PSCBM does not require retraining the full model: it only adds a compact covariance module. This makes it far less computationally demanding than training an SCBM from scratch, while retaining the benefits of modeling concept dependencies. We proposed two training procedures for the covariance module-optimizing the loss without interventions, and optimizing it after intervening on a random subset of concepts. Both variants outperform standard CBMs in test-time accuracy, and under interventions they adapt significantly faster than regular CBMs, though not as rapidly as a fully retrained SCBM. Importantly, training-time interventions improve intervention efficiency without harming accuracy when no interventions are in place, which demonstrates the efficacy of our approach.

Apart from its lightweight nature and efficacy, PSCBM also guarantees compatibility with the original CBM. By disabling the covariance module, PSCBM can revert to identical predictions as the baseline CBM. This property is particularly valuable in regulated domains such as healthcare, where a CBM may already have undergone approval (e.g., FDA testing). In such cases, retraining an SCBM might be prohibited or undermine user trust, whereas PSCBM preserves the validated predictions while still enabling stronger interventions when permitted.

The main limitation of our study is its evaluation on a single dataset; extending the analysis to additional benchmarks will be important for robustness. 
Future work should explore richer training-with-interventions schemes (e.g., varying the number of intervened concepts or using non-random policies), and investigate their applicability in regular SCBMs. Nonetheless, our results show that when retraining from scratch is infeasible, augmenting an existing CBM with a learned covariance matrix provides a simple and efficient way to substancially improve intervention effectiveness.


\section*{Acknowledgements}
SL and MV are supported by the Swiss State Secretariat for Education, Research and Innovation (SERI) under contract number MB22.00047. EP is supported by a fellowship from the ETH AI Center and received funding from the grant \#2021-911 of the Strategic Focal Area “Personalized Health and Related Technologies (PHRT)” of
the ETH Domain (Swiss Federal Institutes of Technology).
\bibliographystyle{iclr2026_conference}
\bibliography{iclr2026_conference}

\newpage

\appendix
\crefalias{section}{appendix}

\section{Related Work}
\label{app:related-work}
\paragraph{Concept Bottleneck Models}
Concept Bottleneck Models (CBM) have originally been proposed by \citet{koh2020concept}. 
The idea is to transform a regular neural network into an interpretable model by converting one of its layers into a concept bottleneck, where each neuron corresponds to a high-level concept that can be interpreted by a human. 
Such a model can be framed as a composite function $f(\vx) = g(h(\vx))$, where $h(\cdot)$ is a concept predictor that calculates the concepts $\hat{\vc}$, and $g(\cdot)$ is the target predictor that predicts the final label from the concepts. 
Since all information that goes from the input to the output passes through the concept bottleneck, the network's prediction should be well explained by the concepts.
\citet{koh2020concept} use what has been called \textit{soft} concept encoding: the target predictor uses real-valued concept logits or probabilities produced by the encoder. However, it has been shown that this method is prone to information leakage, where undesired information can pass through the bottleneck \citep{margeloiu2021do,mahinpei2021promises, makonnen2025measuring}. 
As a remedy, \citet{havasi2022addressing} propose \textit{hard} concepts: the concept encoder predicts concept probabilities, 
which parametrize Bernoulli distributions, from which binary inputs for the target predictor are sampled. 
Additionally, they introduce in their work a side channel 
which allows the target predictor to use information that cannot be expressed in terms of the high-level concepts. This modification can increase target accuracy if the concept set is incomplete.
An alternative approach to modeling the concept bottleneck has been proposed by \citet{espinosa2022concept} who introduce Concept Embedding Models (CEM), where each concept is modeled not by a single neuron but instead by a pair of learnable vector embeddings corresponding to the presence or absence of a concept in the input. This allows passing information not captured by the concepts through the bottleneck without the need for a side channel.
Several methods have been proposed to transform regular models into Concept Bottleneck Models. \citet{yuksekgonul2023post} propose a method to convert one layer of a neural network trained without concepts into a concept bottleneck by using only a small concept-annotated subset. \citet{oikarinen2023label} demonstrate that concepts can also be generated with the help of a Vision-Language Model such as CLIP \citep{radford2021learning}, reducing the human effort associated with annotating data. This work focuses on vision tasks but can be further extended to other domains as applications of CBMs have been explored as well in the context of text~\citep{laguna2025interpretable} or tabular data~\citep{zarlengatabcbm}.
\paragraph{Concept interventions}
A particular advantage of CBMs is the possibility of affecting downstream predictions by modifying incorrectly predicted concepts. This process is called a concept intervention.
\citet{koh2020concept} only consider interventions on randomly selected concepts. On the other hand, \citet{chauhan2023interactive,shin2023closer} propose and evaluate more efficient concept selection policies, such as choosing the concept with the highest associated prediction uncertainty, that is, with probability closest to 0.5. 
This policy allows reducing target prediction error by half after 12 concept interventions on average, while with random concept selection an average of 43 interventions is necessary to achieve it.
In \citet{espinosa2022concept}, a regularization technique called RandInt is introduced, whereby randomly selected concepts are set to their true value during training. This is then shown to increase the model's responsiveness to test-time concept interventions.
As an alternative for an explicit concept selection policy, \citet{espinosa2024learning,singhi2024improvinginterventionefficacyconcept} parameterize it by a neural network which can be trained jointly with the model or in a post-hoc manner.
A further extension of the interventions framework is done by \citet{laguna2024beyond}, introducing a method for intervening on a black-box model without altering its architecture. They also define a measure of intervenability and show that fine-tuning a model for this quantity can improve the performance of interventions.

\paragraph{Modeling concept dependencies}
Since correlated concepts may require multiple interventions, several methods model dependencies. \citet{havasi2022addressing} use an autoregressive structure, by which the value of every concept depends on the previous concepts. 
It increases the model's accuracy, but has the disadvantage that concepts have to be evaluated sequentially, which increases evaluation time.
\citet{singhi2024improvinginterventionefficacyconcept} augment the CBM with a neural network that is trained to update all concepts in response to an intervention made to one of them. This can be combined with the trainable concept selection policy mentioned above. These elements can be either trained jointly with the full model or added to an existing CBM post-hoc. However, unlike our method, it does not provide an \textit{explicit} representation of concept correlations.
Finally, \citet{Vandenhirtz2024stochastic} introduce Stochastic Concept Bottleneck Models (SCBM), in which concept correlations are captured by a multivariate normal distribution. It requires predicting not only the expected value of the concepts, but also a covariance matrix. A special advantage of this model is the explicit representation of concept dependencies, which is used in an intervention strategy based on the distribution's confidence region.

\newpage
\section{Comparing Different Intervention Strategies and Policies}
\label{app-intervention-strategies}

In \Cref{sec:methods}, we decomposed a concept intervention into the selection of concepts for intervention and updating their values. The former part we denote as \textit{intervention policy} and the latter - as \textit{intervention strategy}.
In the following we describe and evaluate different possible choices.
We consider two intervention policies: \textit{random} concept selection (as in \citet{koh2020concept}) and \textit{concept uncertainty} policy, which selects the concept whose predicted probability is closest to 0.5 (that is, the most uncertain concept). 
For intervention strategies we consider four options:
\begin{enumerate}
    \item \textit{Empirical Percentile Strategy} sets the value of the intervened-on concept to the 5th (if it is 0) or 95th (if it is 1) percentile of the empirical distribution of concept predictions made by the model. \citet{koh2020concept} use it for intervening in CBM with soft concept encoding.
    \item \textit{Hard Strategy} introduced by \citet{havasi2022addressing} sets concept probabilities to 0 or 1. In the case of PSCBM, where concepts are represented by their logits, one cannot directly use the logits of 0-1 probabilities, as they are infinite. Instead, we set concepts to the logits of $\varepsilon$ for absent concepts and $1-\varepsilon$ for present concepts, where $\varepsilon$ is a small positive number.
    \item \textit{Simple Percentile Strategy} is a softened version of the hard strategy, where concept probabilities of absent or present are set to 0.05 or 0.95, respectively.
    \item \textit{Confidence Region Strategy} is a strategy that directly uses the concepts' multivariate normal distribution. If a subset $\mathcal{S} \subset\{1,\dots \mathcal{C}\}$ of concepts has been selected for intervention, the updated concept logits $\mathbf{\eta}'_\mathcal{S} $ are calculated by solving the following optimization problem:
        \begin{equation}
        \begin{aligned}
        &\eta'_S = \arg\max_{\eta_S} \log p(\vc_\mathcal{S} | \eta_\mathcal{S}), \\
        &\text{s.t.} -2 \left(\log p(\eta_\mathcal{S} | \mu_\mathcal{S}, \boldsymbol{\Sigma}_{\mathcal{S},\mathcal{S}}) - \log p(\boldsymbol{\mu}_\mathcal{S} | \boldsymbol{\mu}_\mathcal{S}, \boldsymbol{\Sigma}_{\mathcal{S},\mathcal{S}})\right) \leq \chi^2_{d,1-\alpha} \\
        &\quad\quad\eta_i - \mu_i \geq 0 \quad \text{if } c_i = 1, \quad \forall i \in \mathcal{S} \\
        &\quad\quad\eta_i - \mu_i \leq 0 \quad \text{if } c_i = 0, \quad \forall i \in \mathcal{S}
        \end{aligned}  
        \end{equation}
    where $\mathcal{S}$ is the subset of concepts that have been selected for intervention, $\boldsymbol{\mu}_\mathcal{S}$ and $\boldsymbol{\Sigma}_{\mathcal{S},\mathcal{S}} $ are the predicted concept mean and covariance of this concept subset, $c_i$ are the concept values selected by the user, $d$ is the dimensionality of the distribution, i.e. the number of concepts, and $\alpha$ is the requested confidence of the confidence region.
    In this problem we seek to find concept logits $\eta'_\mathcal{S}$ which maximize the log-probability of intervened concept values, and stay within the $\alpha$-confidence region of the predicted distribution, that is, the model's prediction shouldn't be completely disregarded (first constraint). The log-probability of this distribution, after being multiplied by -2, asymptotically follows the $\chi^2$ distribution \citep{silvey1975statistical}, which gives rise to the first inequality.
    The other two inequalities follow from the requirement that the new concept logits shouldn't move away from the ground truth: if $c_i = 1$, $\eta_i$ should increase, and if $c_i = 0$, it should decrease. This last strategy can only be used in the context of SCBM, while the three others can be used by both SCBM and CBM.
    
\end{enumerate}

In \Cref{tab-int-auc-all-models} we report the AUC for concept and target accuracy for all models and all intervention policies and strategies that can be used with them. The strategy based on confidence region cannot be used with a regular CBM, because it requires the covariance matrix.

Interestingly, there is no single best strategy. The \textit{Hard} one typically produces the best or close-to-the-best results for target accuracy, but the \textit{Confidence Region} strategy is often better on concept accuracy, especially with models using an amortized covariance. 
Hence, the choice of the best strategy is not an obvious one, and we encourage the reader to evaluate different policies in relation to the specific data and setup at hand.

\begin{table}[htbp]
  \centering
  \caption{
  Comparison of AUC for all combinations of strategy and policy pairs for all baseline and PSCBM models. The values are normalized to lie in the interval $[0,1]$. For each model, policy pair, the best value is \textbf{bold}, and those lying within a standard deviation are \underline{underlined}. Likewise, the selected strategy is also \textbf{bold} and the one following up is \underline{underlined}.}
  \label{tab-int-auc-all-models}
  \resizebox{\textwidth}{!}{%
\begin{tabular}{@{}lllcc@{}}
    \toprule
    \textbf{Method} & \textbf{Policy} & \textbf{Strategy} & \textbf{Target Accuracy AUC} & \textbf{Concept Accuracy AUC} \\
\midrule
\multirow{6}{*}{CBM} &
\multirow{3}{*}{Concept Uncertainty} &
\textbf{{Hard}}                      &   $ \mathbf{{0.9551}} \pm \num { 0.0000 } $   &   $ \mathbf{{0.9825}} \pm \num { 0.0000 } $ \\
&&  Simple Percentile                    &   $ 0.9513 \pm \num { 0.0000 } $   &   $ \underline{{0.9825}} \pm \num { 0.0000 } $ \\
&&  Empirical Percentile                 &   $ 0.9550 \pm \num { 0.0000 } $   &   $ 0.9798 \pm \num { 0.0002 } $ \\
\cmidrule{2-5}
&
\multirow{3}{*}{Random} &
\textbf{{Hard}}                      &   $ \mathbf{{0.8927}} \pm \num { 0.0003 } $   &   $ \underline{{0.9658}} \pm \num { 0.0000 } $ \\
&&  Simple Percentile                    &   $ 0.8847 \pm \num { 0.0015 } $   &   $ \mathbf{{0.9659}} \pm \num { 0.0001 } $ \\
&&  \underline{{Empirical Percentile}}   &   $ \underline{{0.8927}} \pm \num { 0.0007 } $   &   $ 0.9642 \pm \num { 0.0001 } $ \\
\midrule
\multirow{8}{*}{SCBM (global covariance)} &
\multirow{4}{*}{Concept Uncertainty} &
\textbf{{Hard}}                      &   $ \mathbf{{0.9671}} \pm \num { 0.0001 } $   &   $ \mathbf{{0.9870}} \pm \num { 0.0000 } $ \\
&&  Simple Percentile                    &   $ 0.9453 \pm \num { 0.0056 } $   &   $ 0.9835 \pm \num { 0.0002 } $ \\
&&  Empirical Percentile                 &   $ 0.9570 \pm \num { 0.0060 } $   &   $ 0.9864 \pm \num { 0.0001 } $ \\
&&  Confidence Region                    &   $ 0.9630 \pm \num { 0.0023 } $   &   $ 0.9866 \pm \num { 0.0001 } $ \\
\cmidrule{2-5}
&
\multirow{4}{*}{Random} &
\textbf{{Hard}}                      &   $ \mathbf{{0.9085}} \pm \num { 0.0010 } $   &   $ 0.9697 \pm \num { 0.0001 } $ \\
&&  Simple Percentile                    &   $ 0.8390 \pm \num { 0.0029 } $   &   $ 0.9147 \pm \num { 0.0002 } $ \\
&&  Empirical Percentile                 &   $ 0.9013 \pm \num { 0.0041 } $   &   $ 0.9637 \pm \num { 0.0004 } $ \\
&&  Confidence Region                    &   $ 0.8963 \pm \num { 0.0014 } $   &   $ \mathbf{{0.9718}} \pm \num { 0.0004 } $ \\
\midrule
\multirow{8}{*}{SCBM (amortized covariance)} &
\multirow{4}{*}{Concept Uncertainty} &
Hard                                 &   $ 0.9646 \pm \num { 0.0000 } $   &   $ 0.9860 \pm \num { 0.0000 } $ \\
&&  Simple Percentile                    &   $ 0.9653 \pm \num { 0.0001 } $   &   $ \mathbf{{0.9873}} \pm \num { 0.0000 } $ \\
&&  Empirical Percentile                 &   $ 0.9653 \pm \num { 0.0002 } $   &   $ 0.9863 \pm \num { 0.0000 } $ \\
&&  \textbf{{Confidence Region}}         &   $ \mathbf{{0.9664}} \pm \num { 0.0001 } $   &   $ 0.9871 \pm \num { 0.0000 } $ \\
\cmidrule{2-5}
&
\multirow{4}{*}{Random} &
\textbf{{Hard}}                      &   $ \mathbf{{0.9234}} \pm \num { 0.0005 } $   &   $ 0.9785 \pm \num { 0.0000 } $ \\
&&  Simple Percentile                    &   $ 0.9099 \pm \num { 0.0019 } $   &   $ 0.9509 \pm \num { 0.0005 } $ \\
&&  \underline{{Empirical Percentile}}   &   $ \underline{{0.9230}} \pm \num { 0.0007 } $   &   $ 0.9782 \pm \num { 0.0001 } $ \\
&&  Confidence Region                    &   $ 0.9204 \pm \num { 0.0008 } $   &   $ \mathbf{{0.9789}} \pm \num { 0.0001 } $ \\
\midrule
\multirow{8}{*}{PSCBM (global covariance)} &
\multirow{4}{*}{Concept Uncertainty} &
\textbf{{Hard}}                      &   $ \mathbf{{0.9680}} \pm \num { 0.0003 } $   &   $ \mathbf{{0.9859}} \pm \num { 0.0001 } $ \\
&&  \underline{{Simple Percentile}}      &   $ \underline{{0.9680}} \pm \num { 0.0002 } $   &   $ \underline{{0.9859}} \pm \num { 0.0001 } $ \\
&&  Empirical Percentile                 &   $ 0.9665 \pm \num { 0.0005 } $   &   $ 0.9849 \pm \num { 0.0001 } $ \\
&&  Confidence Region                    &   $ 0.9657 \pm \num { 0.0002 } $   &   $ 0.9853 \pm \num { 0.0001 } $ \\
\cmidrule{2-5}
&
\multirow{4}{*}{Random} &
\textbf{{Hard}}                      &   $ \mathbf{{0.9081}} \pm \num { 0.0014 } $   &   $ 0.9705 \pm \num { 0.0002 } $ \\
&&  \underline{{Simple Percentile}}      &   $ \underline{{0.9078}} \pm \num { 0.0012 } $   &   $ 0.9705 \pm \num { 0.0002 } $ \\
&&  Empirical Percentile                 &   $ 0.8817 \pm \num { 0.0021 } $   &   $ 0.9648 \pm \num { 0.0004 } $ \\
&&  \underline{{Confidence Region}}      &   $ \underline{{0.9070}} \pm \num { 0.0014 } $   &   $ \mathbf{{0.9716}} \pm \num { 0.0001 } $ \\
\midrule
\multirow{8}{*}{{PSCBM (amortized covariance)}} &
\multirow{4}{*}{Concept Uncertainty} &
\underline{{Hard}}                   &   $ \underline{{0.9665}} \pm \num { 0.0018 } $   &   $ \underline{{0.9859}} \pm \num { 0.0004 } $ \\
&&  \textbf{{Simple Percentile}}         &   $ \mathbf{{0.9665}} \pm \num { 0.0018 } $   &   $ \mathbf{{0.9859}} \pm \num { 0.0004 } $ \\
&&  Empirical Percentile                 &   $ 0.9591 \pm \num { 0.0025 } $   &   $ 0.9832 \pm \num { 0.0007 } $ \\
&&  \underline{{Confidence Region}}      &   $ \underline{{0.9661}} \pm \num { 0.0013 } $   &   $ \underline{{0.9855}} \pm \num { 0.0002 } $ \\
\cmidrule{2-5}
&
\multirow{4}{*}{Random} &
\underline{{Hard}}                   &   $ \underline{{0.9129}} \pm \num { 0.0018 } $   &   $ 0.9638 \pm \num { 0.0046 } $ \\
&&  \underline{{Simple Percentile}}      &   $ \underline{{0.9127}} \pm \num { 0.0016 } $   &   $ 0.9638 \pm \num { 0.0045 } $ \\
&&  Empirical Percentile                 &   $ 0.9046 \pm \num { 0.0017 } $   &   $ 0.9711 \pm \num { 0.0008 } $ \\
&&  \textbf{{Confidence Region}}         &   $ \mathbf{{0.9140}} \pm \num { 0.0011 } $   &   $ \mathbf{{0.9744}} \pm \num { 0.0004 } $ \\
\midrule
\multirow{8}{*}{{PSCBMi (global covariance)}} &
\multirow{4}{*}{Concept Uncertainty} &
\underline{{Hard}}                   &   $ \underline{{0.9708}} \pm \num { 0.0001 } $   &   $ \mathbf{{0.9857}} \pm \num { 0.0000 } $ \\
&&  \textbf{{Simple Percentile}}         &   $ \mathbf{{0.9708}} \pm \num { 0.0002 } $   &   $ \underline{{0.9857}} \pm \num { 0.0000 } $ \\
&&  Empirical Percentile                 &   $ 0.9036 \pm \num { 0.0006 } $   &   $ 0.9666 \pm \num { 0.0001 } $ \\
&&  Confidence Region                    &   $ 0.9379 \pm \num { 0.0003 } $   &   $ 0.9767 \pm \num { 0.0000 } $ \\
\cmidrule{2-5}
&
\multirow{4}{*}{Random} &
\underline{{Hard}}                   &   $ \underline{{0.8846}} \pm \num { 0.0014 } $   &   $ \mathbf{{0.9552}} \pm \num { 0.0003 } $ \\
&&  \textbf{{Simple Percentile}}         &   $ \mathbf{{0.8851}} \pm \num { 0.0007 } $   &   $ \underline{{0.9552}} \pm \num { 0.0002 } $ \\
&&  Empirical Percentile                 &   $ 0.7797 \pm \num { 0.0021 } $   &   $ 0.9321 \pm \num { 0.0002 } $ \\
&&  Confidence Region                    &   $ 0.8321 \pm \num { 0.0013 } $   &   $ 0.9475 \pm \num { 0.0001 } $ \\
\bottomrule \end{tabular}}
\end{table}



\newpage
\section{Pseudocode for Interventions Training}
\label{app-pseudocode}
In \Cref{alg:intervention-loss} we present the pseudocode for one iteration of the intervention-aware training procedure.

\begin{algorithm}
\caption{Loss calculation for one iteration of training with random interventions.}
\label{alg:intervention-loss}
\begin{algorithmic}[1]
\STATE \textbf{Inputs:}
\STATE \quad $g$ (target predictor)
\STATE \quad $\mathcal{L}$ (loss function)
\STATE \quad $L$ (number of concepts for intervention)
\STATE \quad $N$ (number of random interventions per datapoint)
\STATE \quad $\tau$ (intervention strategy, which modifies concept values)
\STATE \quad $(\vc,y)$ (concept and target labels)
\STATE \quad $(\hat{\boldsymbol{\mu}}, \hat{\boldsymbol{\Sigma}})$ (predicted expected value and covariance of concepts)
\STATE \textbf{Algorithm:}

\STATE \quad $l = 0$ (Initialize the loss to 0)
\FOR {i in 1\dots N}
\STATE \quad $S = \text{Random}(L)$ (Randomly select $L$ concepts for intervention)
\STATE \quad ${\boldsymbol{\mu}}', {\boldsymbol{\Sigma}}' = \tau(S,\hat{\boldsymbol{\mu}}, \hat{\boldsymbol{\Sigma}})$  (Use strategy $\tau$ to update the concept distribution)
\STATE \quad $\eta'\sim\mathcal{N}({\boldsymbol{\mu}}', {\boldsymbol{\Sigma}}')$ (Sample concept logits from the updated distribution)
\STATE \quad $\vc' = \sigma(\eta')$ (Pass concept logits through a sigmoid to obtain a vector of probabilities)
\STATE \quad ${\vc}_{1:M} = \text{Bern}(\vc')$ (Sample $M$ concept vectors from the Bernoulli distribution defined by $\vc'$)
\STATE \quad $y'=\frac{1}{M}\sum_{m=1}^M g(\vc^{(m)})$ (Calculate the target)
\STATE \quad  $l = l+\mathcal{L}(y',y,\vc',\vc)$ (Loss after intervention)
\ENDFOR
\quad \RETURN $\frac{l}{N}$ (Average loss)
\end{algorithmic}
\end{algorithm}

\newpage
\section{Results with amortized covariance}
\label{app-amortized-covariance}
\begin{figure}[h!]
    \centering
    \includegraphics[width=\linewidth]{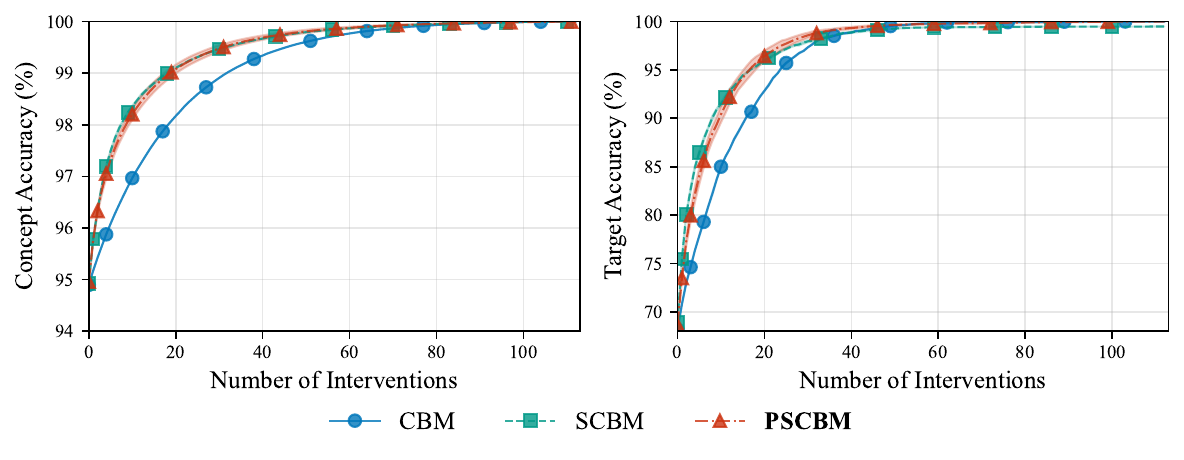}
    \caption{Concept and Target Accuracy for amortized PSCBM models and baselines when Concept Uncertainty Policy is used. The shadings represent the standard deviation.}
    \label{fig:prob_unc-local-models}
\end{figure}

In this appendix, we show additional results for the studied models using an \textit{amortized} covariance matrix instead of a globally learnt one as in the main manuscript. The amortized covariance, as introduced in SCBM~\cite{Vandenhirtz2024stochastic}, makes the $\boldsymbol{\mu}$ and $\boldsymbol{\Sigma}$ learnt per instance, i.e. amortized by the input $\vx$. We show results for CBM and SCBM as baselines and for a PSCBM trained without interventions.
In \Cref{tab-test-accuracy-amortized} we show test-time results for target and concept accuracy, as well as target and concept accuracy AUC for interventions with probability uncertainty policy.
In \Cref{fig:prob_unc-local-models}, we show the respective intervention curves. We observe that without interventions, the performance of PSCBM is better than thus of the baselines. With interventions, it scores the highest on target accuracy AUC, and on par with the rest on concept accuracy AUC.
Intervention curves confirm these results. However, in the curve for target accuracy, one can observe that after the first few interventions, the SCBM improved faster than the PSCBM. This is important because in a real-world scenario it is expected that only a few concept interventions would be made.

\begin{table}[h!]
    \centering
    \caption{Test-time target and concept accuracy without interventions and their AUCs during interventions in PSCBM and baselines with an \textit{amortized} covariance matrix. AUC is normalised to stay within the interval [0,1]. We report the mean and standard deviation over three runs. The best scores for each metric are emphasized with \textbf{bold} font, and those within a standard deviation from the best one are \underline{underlined}.}
    \label{tab-test-accuracy-amortized}
    \resizebox{\textwidth}{!}{%
    \begin{tabular}{@{}lccccc@{}}
    \toprule
    \textbf{Method} & \textbf{Target Accuracy (\%)} & \textbf{Concept Accuracy (\%)} & \textbf{Target Accuracy AUC} & \textbf{Concept Accuracy AUC}  & \textbf{Training time (s)}\\
    CBM & $ 67.3973 \pm 0.5722 $ & $ \underline{94.9403} \pm 0.1059 $ & $ \underline{0.9551} \pm 0.0000 $ & $ 0.9825 \pm 0.0000 $ & $7204  \pm        247  $ \\  
    SCBM & $ \underline{68.6342} \pm 0.3790 $ & $ \underline{94.9190} \pm 0.0074 $ & $ 0.9646 \pm 0.0000 $ & $ \mathbf{0.9860} \pm 0.0000 $ & $ 8130  \pm       1073  $\\
    PSCBM & $ \mathbf{68.6611} \pm 0.2067 $ & $ \mathbf{94.9605} \pm 0.0639 $ & $ \mathbf{0.9665} \pm 0.0018 $ & $ \underline{0.9859} \pm 0.0004 $ & $   670  \pm  55  $ \\
    \bottomrule
    \end{tabular}}
\end{table}

\newpage

\section{Implementation Details}
\label{app-implementation}
All models have been implemented in PyTorch \citep{ansel2024pytorch} and optimized using Adam \citep{kingma2015adam}. To apply weight decay, we used the optimizer AdamW, which decouples weight decay from gradient computation \citep{loshchilov2019decoupledweightdecayregularization}. All models use a ResNet-18 encoder. The remaninig parts including the concept mean and covariances, and the target predictor are all linear layers.
All baselines have been trained with three random initializations. All PSCBM versions have been evaluated with each of the three CBMs with three random seeds, which amounts to a total number of nine evaluations. The computations have been done on a cluster composed mostly of NVIDIA GeForce RTX 2080 GPUs, and every job used a single GPU and two CPUs.
When we train PSCBM with interventions, we use 20 random masks per datapoint per epoch, and found that a larger number does not lead to better performance. We intervene on 20\% of the total number of concepts, and their values at training time are set according to the \textit{Hard} Strategy, introduced in \Cref{app-intervention-strategies}.
We performed hyperparameter optimization with respect to learning rate scheduling, initial learning rate, and weight decay factor. 
In \Cref{tab-optimal-hyperparameters} we report the optimal values based on validation loss after the last training epoch. It should be noted that PSCBMa is not very sensitive to weight decay and that for PSCBMg the advantage of step learning rate over cosine schedule was minimal, albeit statistically significant. Similarly, for CBM and for both SCBM variants, all four tested hyperparameter combinations achieved similar performance.


\begin{table}[hbtp]
  \caption{Optimal hyperparameter values for each model.}
  \label{tab-optimal-hyperparameters}
  \centering
  \begin{tabular}{llll}
    \toprule
    Model     & Learning rate scheduler & Initial learning rate & Weight decay \\
    \midrule 
    CBM & Step-wise & $10^{-4}$ & $0.1$ \\
    SCBMg & Step-wise & $10^{-4}$ & $0.1$ \\
    SCBMa & Step-wise & $10^{-4}$ & $0.1$ \\
    PSCBMg & Step-wise & $10^{-3}$ & $1$ \\
    PSCBMa & Cosine & $10^{-4}$ & $0$ \\
    PSCBMg\textsubscript{i} & Cosine & $10^{-4}$ & $4$ \\
    \bottomrule              
  \end{tabular}
\end{table}

\newpage
\section{Results with Random Policy}
\label{app-random-policy}
\begin{figure}[h!]
    \centering
    \includegraphics[width=\linewidth]{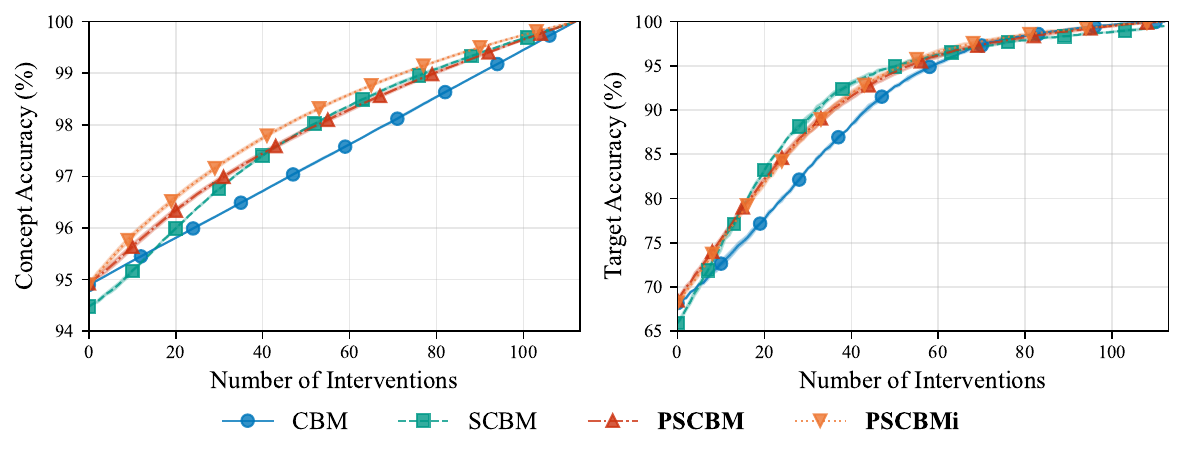}
    \caption{Concept and Target Accuracy for PSCBM models and baselines when Random policy is used. The shadings represent the standard deviation.}
    \label{fig:random-all-models}
\end{figure}

In \Cref{fig:random-all-models} we show the intervention curves when concepts for intervention are chosen randomly. One can observe generally lower improvement rates in comparison to the selection policy based on concept uncertainty. 
On concept accuracy, both PSCBM variants outperform SCBM. On target accuracy, they outperform the regular CBM but  fall slightly behind the SCBM.

\end{document}